\documentclass[11pt,a4paper]{article}
\usepackage[hyperref]{acl2020}
\usepackage{times}
\usepackage{latexsym}

\usepackage{microtype}
\usepackage{graphicx}

\aclfinalcopy 

\title{FH-SWF\_SG at GermEval 2021: Using Transformer-Based Language Models to Identify Toxic, Engaging, \& Fact-Claiming Comments}

\author{Christian Gawron \\
  Fachhochschule S\"udwestfalen \\
  Frauenstuhlweg 31 \\
  58644 Iserlohn \\
  \texttt{gawron.christian@fh-swf.de} \\\And
  Sebastian Schmidt \\
  Fachhochschule S\"udwestfalen \\
  Frauenstuhlweg 31 \\
  58644 Iserlohn \\
  \texttt{schmidt.sebastian2@fh-swf.de} \\}

\date{}

\begin{document}
\maketitle
\begin{abstract}
    In this paper we describe the methods we used for our submissions to the 
    GermEval 2021 shared task on the identification of toxic, engaging, and fact-claiming comments.
    For all three subtasks we fine-tuned freely available transformer-based models from the 
    Huggingface model hub. We evaluated the performance of various pre-trained models 
    after fine-tuning on 80\% of the training data
    with different hyperparameters and submitted predictions of the two best performing resulting models.
    We found that this approach worked best for subtask 3, for which we achieved an F1-score of $0.736$. 
\end{abstract}

\section{Introduction}
Compared to the detection of offensive language in GermEval 2018 \cite{GermEval2018} and 2019 \cite{GermEval2019}, 
this year's task adds two important additional categories found in social media comments, namely \emph{engaging} and 
\emph{fact-claiming} comments \cite{germeval2021overview}. With federal elections being held in 2021, identifying 
fact-claiming statements (subtask 3) in German social media posts has gained additional relevance as ``fake news'' might have 
had an influence on other important elections, e.\,g. the 2016 US presidential elections \cite{10.1257/jep.31.2.211, Bovet2019}.
A system identifying fact-claiming comments could help to identify potential attempts to spread false factual statements. 

The identification of \emph{engaging} comments (subtask 2) is potentially interesting for the 
ranking algorithms used by social network providers. 
Increasing the visibility of these comments might help improving the attractiveness of a social network 
by encouraging the users to employ a more respectful and rational style of discussion.

With the classification of toxic comments (subtask 1), the GermEval Shared Tasks on the identification of 
offensive language mentioned above are continued. This category is also useful for the ranking algorithms
of social media providers and could be used to decrease the visibility of such comments.
However, we have made the experience that this year's \emph{toxic} category is harder to identify than the former 
offensive categories -- at least by our approach. 

The best performing systems in GermEval 2019 were based on BERT \cite{BERT}. 
Leveraging the transformer architecture \cite{transformer} with its attention
mechanism, BERT is able to model relations between words and to create semantic embeddings of sentences 
\cite{DBLP:journals/corr/abs-2007-01852}.  
In the last two years, various modifications of BERT 
like RoBERTa \cite{RoBERTa} or ELECTRA \cite{ELECTRA} have been proposed and shown to achieve state-of-the-art results on 
various NLP tasks. 
Other transformer-based models,
especially GPT-2 \cite{GPT2} and its successor GPT-3, even made it into the press \cite{Zeit_GPT3} due to their ability to create 
high-quality artificial text or to create source code for various programming languages \cite{GPT3-Code}. 

Probably the most important feature of these models is that they allow transfer learning: After an unsupervised \emph{pre-training}, 
the resulting models can be \emph{fine-tuned} for various NLP tasks like token classification (e.\,g. NER) and sequence classification.
Pre-training a language model for German imposes two challenges: It requires a large corpus of text and is computationally expensive.
According to \citet{GPT3}, GPT-3 was trained on a corpus of 400 billion byte-pair-encoded tokens or roughly 570~GB of text.
Compared to this, the ``Huge German Corpus''\footnote{See \url{https://www.ims.uni-stuttgart.de/forschung/ressourcen/korpora/hgc}} with 204 million tokens is 
rather small. 
BERT-large was trained on 64~TPU chips for four days at an estimated cost of \$7,000 \cite{10.1145/3381831}, 
the training of GPT-3 took 3.640 petaflop-days \cite{GPT3}.
Due to the high computational effort and costs to train a model from scratch, we decided to evaluate freely available pre-trained models for our system.

For English, pre-trained models of high quality are freely available for most of the model architectures mentioned above (with the notable exception of GPT-3).
Unfortunately, the groups which developed and trained these models and the companies behind them do not deem German 
important enough to provide pre-trained models for German. 
Although there is currently no active academic community in Germany training and publishing these language models, 
there is a growing number of companies and individuals publishing such pre-trained models.
For example, Deepset.ai has published a German ELECTRA model achieving an F1-score (macro average) of 80.70\% on GermEval 2018 Coarse and 
an F1-score (micro average) of 88.95\% on GermEval 2014 \cite{GNLM}.
Philipp Reissel and Philip May have published both a 
German ELECTRA model \cite{german-nlp-group/electra-base-german-uncased}
and a ``German colossal, cleaned Common Crawl corpus'' (GC4) \cite{GC4} with about 540 GB of German text 
from the web
It would be helpful for the development of language models for German if an extensive and high-quality corpus of German language text would be available 
through infrastructure projects like CLARIN-D \cite{ClarinD}.

\section{Setup}
Our experiments were performed using Jupiter Notebooks \cite{Kluyver2016jupyter}.
This had the advantage that we could use local computing resources and cloud platforms 
like Google Colaboratory \cite{Bisong2019} without modifications to the code.
The code used to generate our submissions is available on 
GitHub\footnote{The repository \url{https://github.com/fhswf/GermEval2021} will be made public after the 
submission of this paper.}.

We used the web application \emph{Weights \& Biases} \cite{wandb} to record and compare the results of experiments with 
different language models and hyperparameters (learning rate, number of training epochs), which was of great help especially 
when using cloud-based computing resources without a persistent storage medium.

\section{Model Library}
A large repository of pre-trained transformer based language models along with an open-source library of implementations 
of them is operated by Huggingface \cite{huggingface}.
As of July 2021, the \emph{model hub} contains about 2,900 pre-trained models for English and more than 200 pre-trained 
models for German provided by a fast-growing number of contributors, including the groups mentioned above.
Due to the large number of available pre-trained models for German, we decided to use the Huggingface transfer library 
for our submission and to choose among the models available on the model hub.

The transformer library makes it very easy to use and to fine-tune the models provided on the hub. 
Besides the model implementations, it also contains recent optimization algorithms like AdamW \cite{AdamW} and 
Adafactor \cite{Adafactor}, provides integration with the experiment-tracking software \emph{Weights \& Biases} \cite{wandb},
code for loading and handling training data, and commonly used metrics.

\begin{table*}[t]
  \begin{tabular}{lrrrrrrrrr}
    & \multicolumn{3}{c}{Sub1\_Toxic} & \multicolumn{3}{c}{Sub2\_Engaging} & \multicolumn{3}{c}{Sub3\_FactClaiming} \\ 
    Submission & \multicolumn{1}{c}{F1} & \multicolumn{1}{c}{Prec.} & \multicolumn{1}{c}{Rec.} & 
                 \multicolumn{1}{c}{F1} & \multicolumn{1}{c}{Prec.} & \multicolumn{1}{c}{Rec.} & 
                 \multicolumn{1}{c}{F1} & \multicolumn{1}{c}{Prec.} & \multicolumn{1}{c}{Rec.} \\
    \hline
    deepset/gelectra-large & \textbf{0.707} & 0.743 & 0.675 & \textbf{0.697} & 0.694 & 0.700 & 0.734 & 0.728 & 0.740 \\
    benjamin/gerpt2-large & 0.658 & 0.678 & 0.640 & 0.690 & 0.684 & 0.696 & \textbf{0.736} & 0.736 & 0.735
  \end{tabular}
  \caption{\label{tab:results}Results of our submissions based on the models deepset/gelectra-large and benjamin/gerpt2-large.}
\end{table*}

\section{Data Preprocessing}
The transformer-based language models we used for our experiments use either SentencePiece \cite{SentencePiece} or 
byte pair encoding \cite{bpe} for tokenization and can handle rare words and emojis. So we did actually not preprocess the texts in any way.

One of the models we used in our experiments, \texttt{german-nlp-group/electra-base-german-uncased}, is an uncased model that converts all characters 
to lower case during tokenization. Unlike other `uncased' models published on the model hub, this model does not remove accents.

\section{Model Selection}
\label{model-selection}
\begin{figure}[h]
  \includegraphics[width=\linewidth]{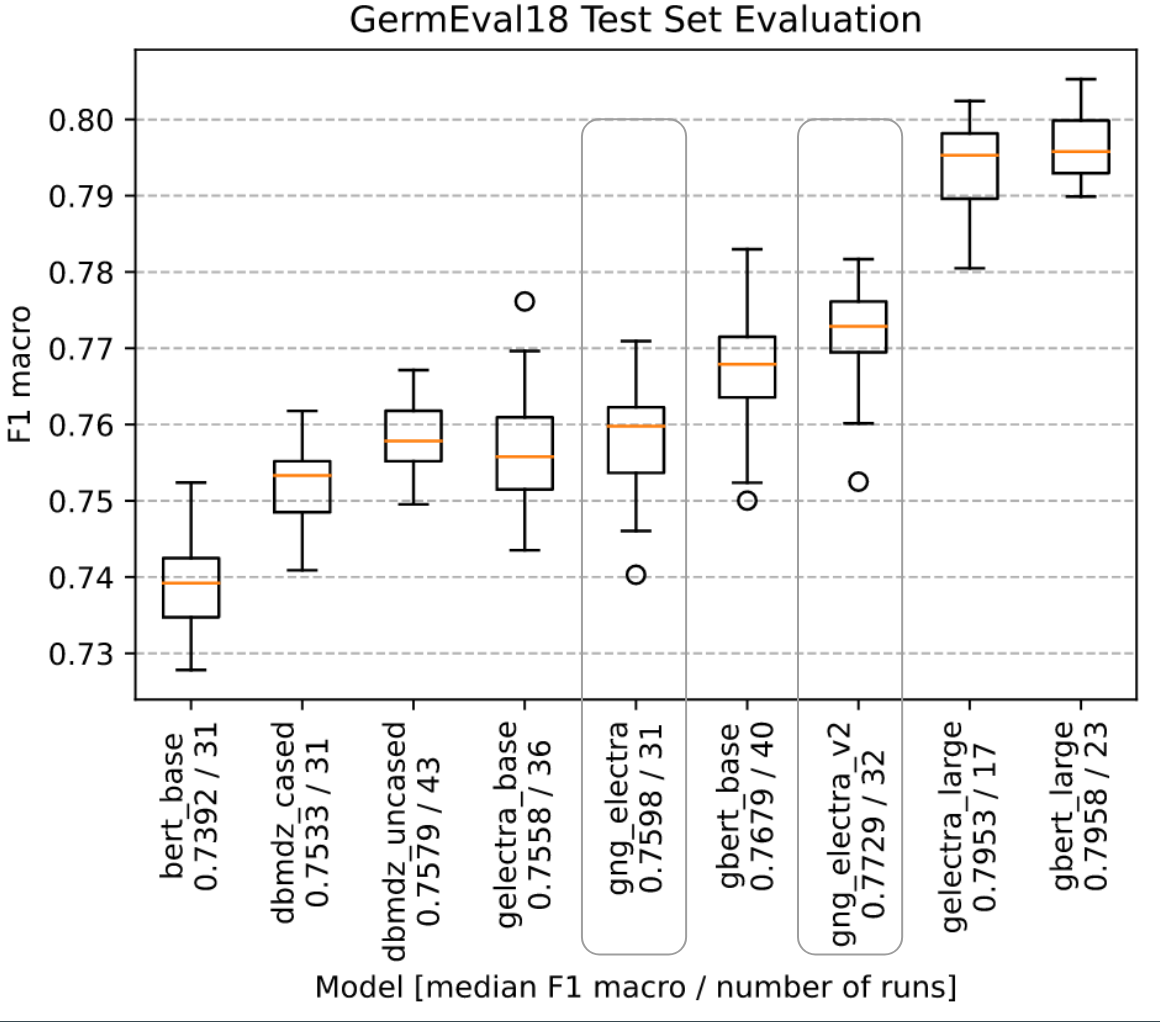}
  \caption{ \label{model-eval} Results of some German language models on the GermEval 2018 dataset. Figure by Philip May, taken from the 
  \texttt{german-nlp-group/electra-base-german-uncased} model card.}
\end{figure}

With more than 200 pre-trained models for German available on the model hub, we needed to do some 
preselection for our experiments. Philip May, one of the authors of \texttt{german-nlp-group/electra-base-german-uncased}, has evaluated several 
models on the GermEval 2018 dataset (see figure \ref{model-eval}).

We chose the best three models of this evaluation as our candidates.
Due to the success of GPT-2 on various NLP tasks \cite{GPT2}, we also included  
\texttt{benjamin/gerpt2-large}, a German GPT-2 model recently published by \citet{benjamin/gerpt2-large}, 
an AI student from Johannes Kepler Universit\"at Linz.

The following list contains some information on these models. Since we are not sure how to calculate the number of model parameters from the specification in the model 
configuration file, we specify the size of the binary file containing the model parameters as a measure of model complexity.

\begin{description}
  \item[\texttt{gbert-large}] has been published by \citet{GNLM}. It is a large BERT model with a binary size of 1.3 GB.
  \item[\texttt{gelectra-large}] by the same group is a German ELECTRA model. The binary size is also 1.3 GB.
  \item[\texttt{electra-base-german-uncased}] by \citet{german-nlp-group/electra-base-german-uncased} is a smaller ELECTRA model with a binary size of 424 MB.
  \item[\texttt{gerpt2-large}] published by \citet{benjamin/gerpt2-large} is a GPT-2 model using an embedding dimension of 1280, 1024 position encodings and 20 attention heads. Although 
  GPT-2 is mainly used for text generation, it also produces sentence embeddings which can be used for text classification. The transformer library 
  provides the class \texttt{GPT2ForSequenceClassification} for this purpose. With a size of 3.2 GB it is the largest model we used.
\end{description}

\section{Computing Resources}
Most calculations were done on a local server using a Tesla V100S GPU card. We used fp16 precision for the training runs on the V100S for better 
performance as some tests with double precision did not show better results.
In addition, we used cloud-based computing resources provided by GraphCore and Google Colaboratory.

\begin{figure}[h]
  \includegraphics[width=\linewidth]{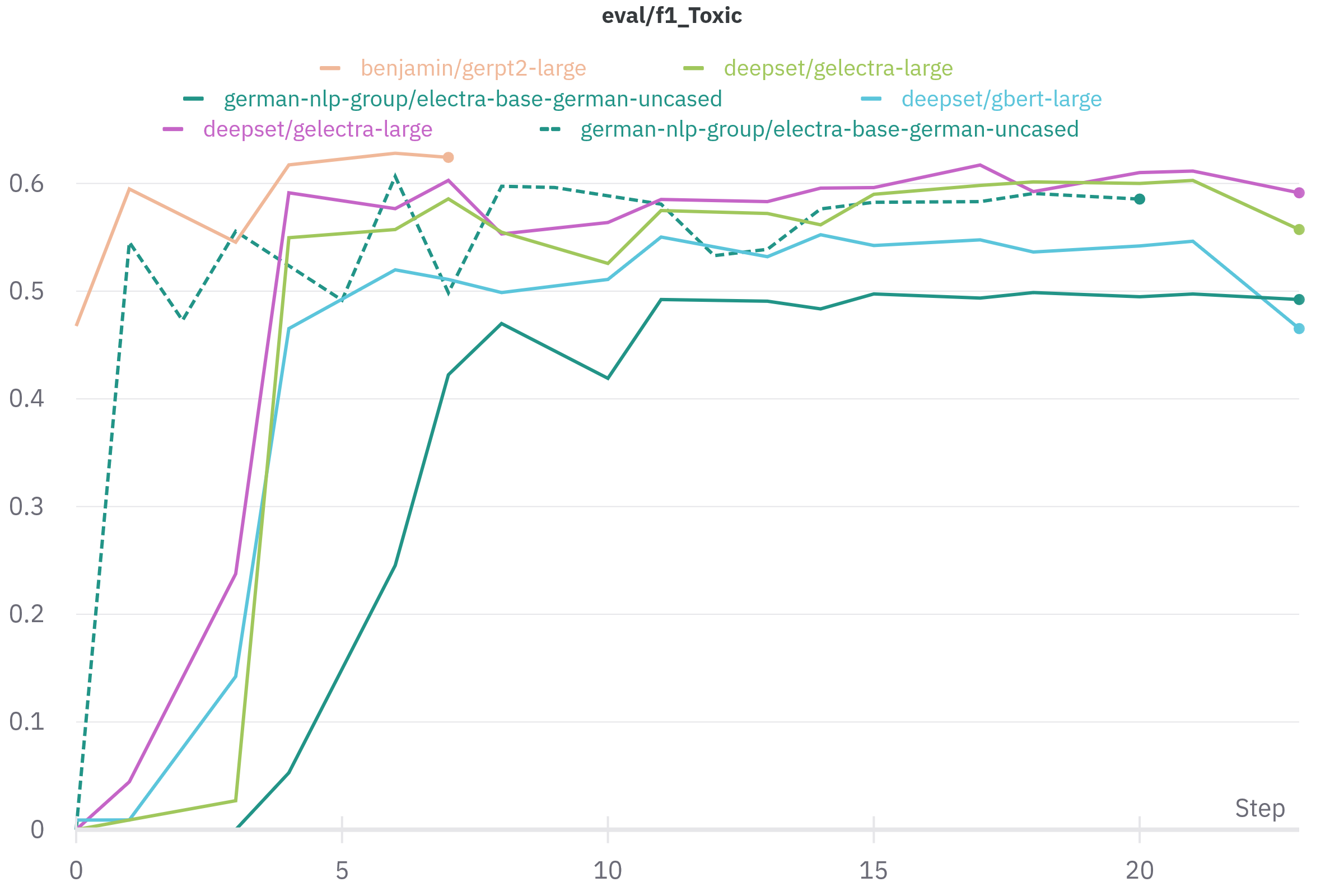}
  \caption{\label{plot:F1_Sub1}F1 scores of different experiments for subtask 1 with a train-test split of 0.8.}
\end{figure}

\begin{figure}[h]
  \includegraphics[width=\linewidth]{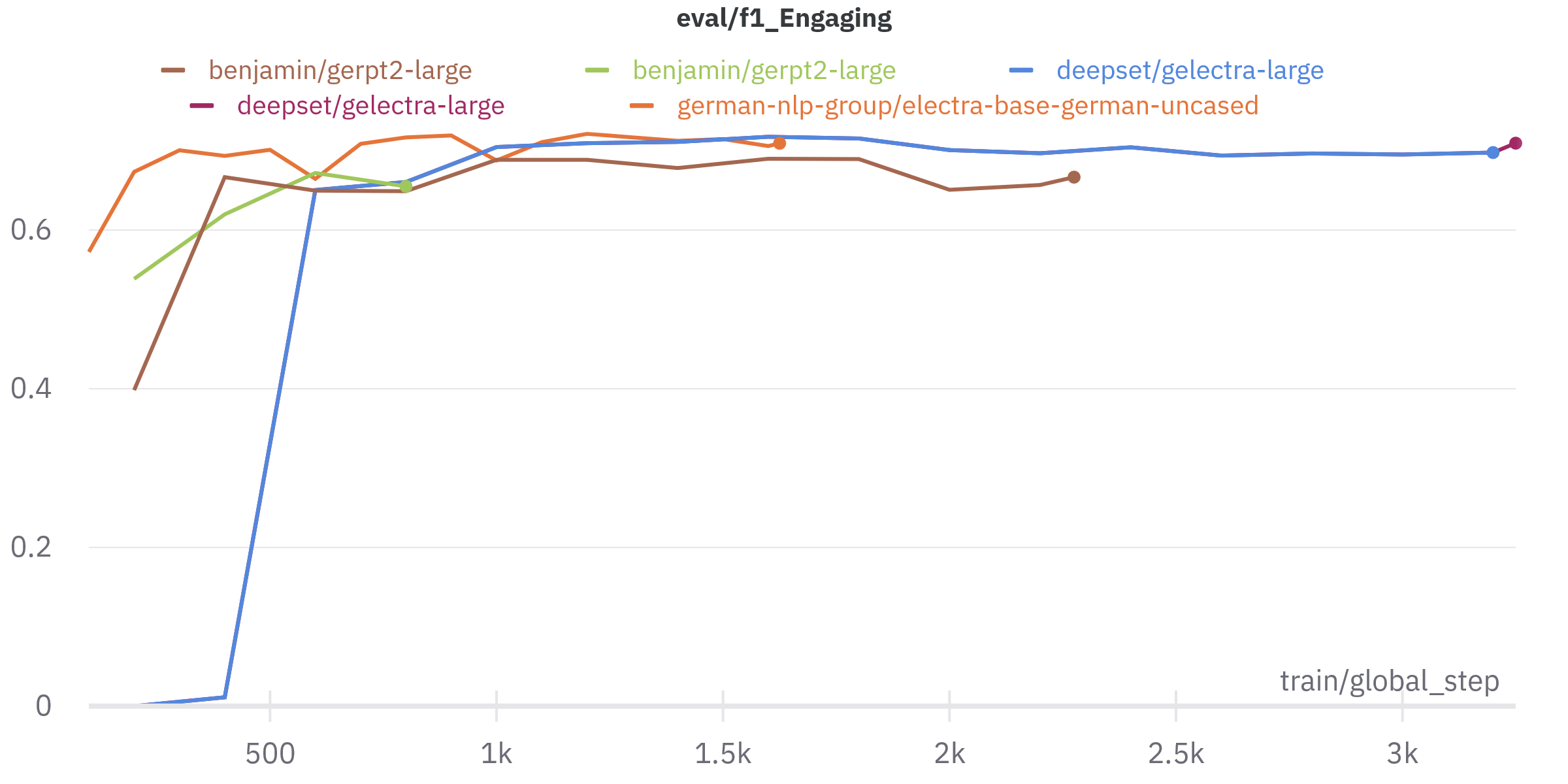}
  \caption{\label{plot:F1_Sub2}F1 scores of different experiments for subtask 2 with a train-test split of 0.8.}
\end{figure}

\begin{figure}[h]
  \includegraphics[width=\linewidth]{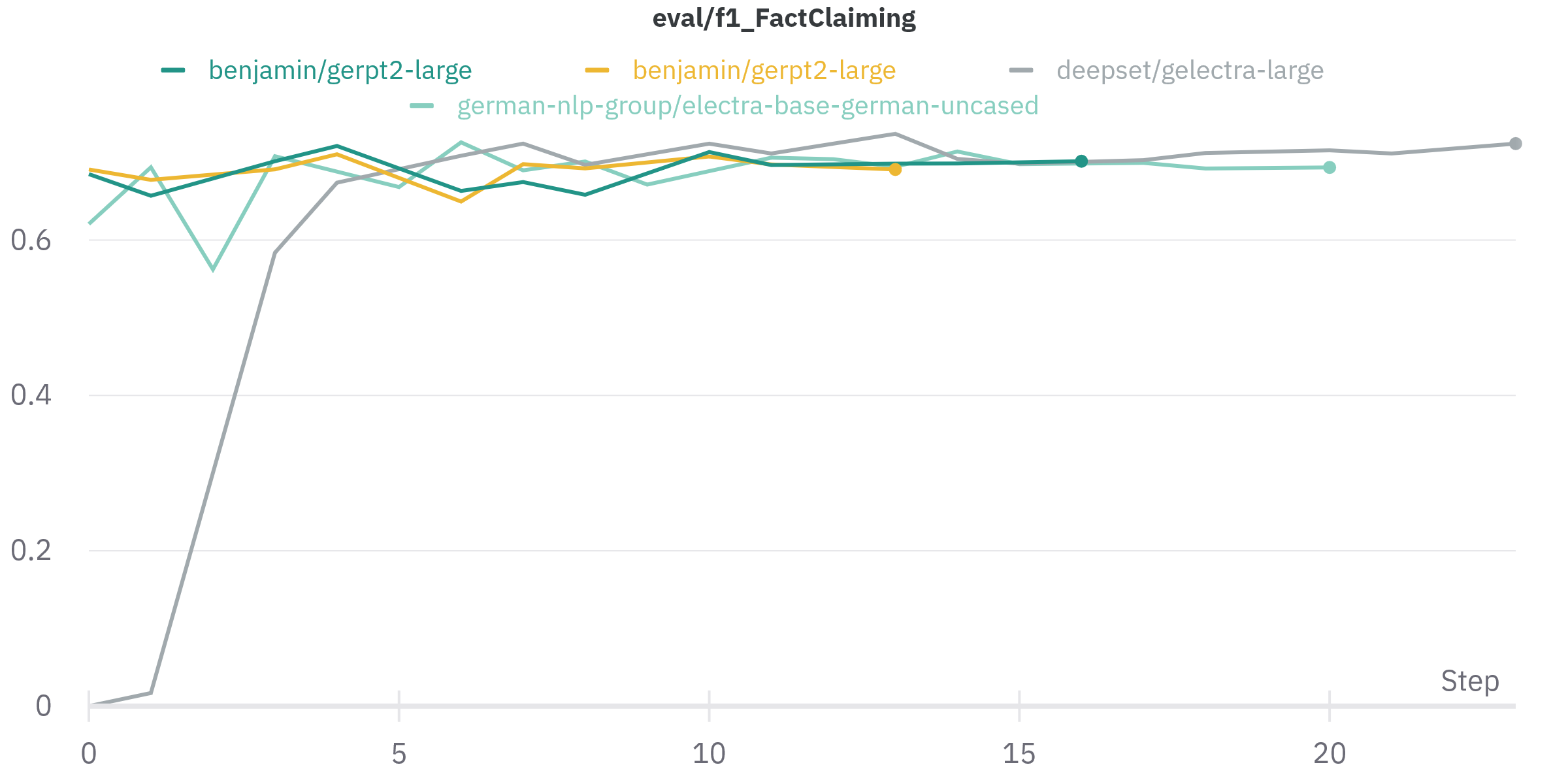}
  \caption{\label{plot:F1_Sub3}F1 scores of different experiments for subtask 3 with a train-test split of 0.8.}
\end{figure}

\section{Results}
Using the four models (see section \ref{model-selection}) we performed several training runs with a train-test split of 80\%.
We did not have the time and computing resources to do a systematic hyperparameter optimization but rather tried different 
learning rates and number of training epochs. 
Figures \ref{plot:F1_Sub1} -- \ref{plot:F1_Sub3} show the resulting F1-scores of several runs ans models for the three subtasks.
Unfortunately, the fluctuations of the F1-scores measured on the 20\% test split during the training were about as large as 
the differences between the different models. At this point, we would have needed more time and resources to perform a larger 
number of training runs and a statistical analysis similar to the one shown in figure \ref{model-eval}. 
In some runs, declining F1-scores at the end of the training runs indicated overfitting -- additional training data would 
probably have improved the results.

Overall, we achieved the best results by fine-tuning \texttt{deepset/gelectra-large} and \texttt{bjamin/gerpt2-large}.
For the final system submissions, we fine-tuned these two models using the complete training dataset for all three subtasks.
Table \ref{tab:results} shows the scores of the two submissions on the test data of the Shared Task.

\section{Using Additional Training Data}
Assuming that offensive language is also considered toxic,
we tried to add data from GermEval 2018 and 2019 to our training dataset for subtask 1.
However, compared to experiments without this additional training data, accuracy and F1-score on our 
validation dataset (i.\,e. 20\% of this year's training data)
were worse for these experiments. 
At least for an AI, toxic comments on facebook seem to be quite different from offensive language 
used on twitter.

\section{Error Analysis}
Before the gold labels were released, we compared our model predictions with our personal predictions for the first test comments.
When we looked at the gold labels, we were surprised by some of the labels, especially with respect to examples having more than one label.

For example, our system flagged a fact claim in comment 3246
\begin{quotation}
  @USER , ich glaube,Sie verkrnnen gr\"undlich die Situation. Deutschland mischt sich nicht ein, weil die letzte Einmischung in der Ukraine noch 
  nicht bereinigt ist. Es geht nicht ums Milit\"ar
\end{quotation}
which we considered correct. We did not expect that this comment is also considered engaging.

In the case of comment 3248
\begin{quotation}
  Als jemand, der im real existierenden Sozialismus aufgewachsen ist, kann ich \"uber George Weineberg nur sagen, dass er ein Voll...t ist. 
  Finde es schon gut, dass der eingeladen wurde. Hat gezeigt, dass er viel Meinung hat, aber offensichtlich wenig Ahnung. 
  Er hat sich eben so gut wie er kann, f\"ur alle sichtbar, zum Trottel gemacht.
\end{quotation}
we agreed with our system that the second sentence (``I think it's good that he was invited'') could be considered engaging, but according to
the gold labels, this comment is only toxic.
On the other hand, comment 3269
\begin{quotation}
  Sry aber Preetz hat nicht viel beizutragen. Er MUSS der Politik in den Hintern kriechen damit sein Verein Zuschauer ins Stadion bekommt. 
  Er ist abh\"angig von der Politik.
\end{quotation}
is both toxic and engaging according to the gold labels, while we agreed with our system that this is only toxic.

These three examples demonstrate that this year's task is really hard -- even for humans.
It would be interesting to measure the score of human annotators getting just the category names and the training examples. 

\section{Conclusion}

When we first looked at the development data, our impression was that fact-claiming statements would be the 
hardest category to recognize for an NLP system due to the wide range of different facts in the statements. 
The rather low range of annotator agreement of $0.73 < \alpha < 0.84$ for subtask 3 also suggests that this 
should be the ``hard'' category. We were quite surprised that our system actually achieved the best F1-score
($0.736$ in the case of \texttt{benjamin/gerpt2-large}) for this category. 

Regarding the toxic category, the F1-score of $0.707$ on subtask 1 is surprisingly low considering the 
F1-score of \texttt{deepset/gelectra-large} of about $0.80$ reported by \citet{GNLM} on 
GermEval 2018 (coarse).  This year's `toxic' category seems to be quite different from the offensive language category 
of the GermEval tasks in 2018 and 2019 and -- at least for an AI -- more difficult to recognize.

The approach we used to create our submissions is a rather simple one that did not require preprocessing of the training 
data or much programming.
Free libraries containing implementations of a wide range of language models and the availability 
of an increasing number of pre-trained model instances make it quite easy to apply state-of-the-art language models for 
NLP tasks like text classification. 
It still, however, requires some coding to train and select models and to create predictions for the test dataset.
Integrated tools like the recently announced 
AutoNLP\footnote{See \url{https://huggingface.co/autonlp}} will probably enable non-experts (and non-coders) to train such 
models in the next few years.

\section*{Acknowledgments}
This research was supported by grants from NVIDIA and utilized NVIDIA CUDA on Tesla \& Ampere GPUs.
This research also used free computing resources provided by the GraphCore Academic Program and Google Colab.

\bibliography{nlp}
\bibliographystyle{acl_natbib}

\end{document}